\documentclass[10pt,twocolumn,letterpaper]{article}

\usepackage{cvpr}            
\usepackage{graphicx}
\usepackage{amsmath}
\usepackage{amssymb}
\usepackage{booktabs}
\usepackage{bm}
\usepackage{multirow}
\usepackage{makecell}

\usepackage[pagebackref,breaklinks,colorlinks]{hyperref}
\usepackage[capitalize]{cleveref}
\crefname{section}{Sec.}{Secs.}
\Crefname{section}{Section}{Sections}
\Crefname{table}{Table}{Tables}
\crefname{table}{Tab.}{Tabs.}

\begin{document}

\title{DreamCom: Finetuning Text-guided Inpainting Model for Image Composition}

\author{Lingxiao Lu, Jiangtong Li, Bo Zhang, Li Niu\thanks{Corresponding author.} \\
Department of Computer Science and Engineering, MoE Key Lab of Artificial Intelligence, \\
Shanghai Jiao Tong University\\
{\tt \small \{lulingxiao,keep\_moving-Lee,bo-zhang,ustcnewly\}@sjtu.edu.cn}
}

\maketitle

\begin{abstract}
The goal of image composition is merging a foreground object into a background image to obtain a realistic composite image. 
Recently, generative composition methods are built on large pretrained diffusion models, due to their unprecedented image generation ability. 
However, they are weak in preserving the foreground object details. 
Inspired by recent text-to-image generation customized for certain object, we propose DreamCom by treating image composition as text-guided image inpainting customized for certain object.
Specifically , we finetune pretrained text-guided image inpainting model based on a few reference images containing the same object, during which the text prompt contains a special token associated with this object. Then, given a new background, we can insert this object into the background with the text prompt containing the special token. 
In practice, the inserted object may be adversely affected by the background, so we propose masked attention mechanisms to avoid negative background interference.
Experimental results on DreamEditBench and our contributed MureCom dataset show the outstanding performance of our DreamCom.
\end{abstract}

\section{Introduction} \label{sec:intro}

Image composition \cite{compositionsurvey} aims to produce a high-quality composite image by combining the foreground and background from different image sources, which has a wide range of applications including artistic creation, E-commerce, and data augmentation. 
Early works \cite{deepblending,deepharmonization,shadow2} decompose image composition into different sub-tasks (\eg, image blending, image harmonization) to solve different issues (\eg, unnatural boundary, inharmonious illumination) in the composite image. 
To obtain a realistic composite image by addressing all the issues, multiple sub-tasks need to be performed sequentially, which is very cumbersome and thus hinders real-world applications. 

Recently, some works \cite{paintbyexample,objectstitch,li2023dreamedit} attempt to solve all the issues with one unified diffusion model.
Given a foreground image and a background image with a bounding box specifying the object placement, they could directly produce the final composite image with the foreground object seamlessly and naturally placed in the bounding box. 
These methods can be roughly categorized into two groups: object-to-object mapping and token-to-object mapping.

The first group (\eg, PbE~\cite{paintbyexample}, ObjectStitch~\cite{objectstitch}, ControlCom~\cite{zhang2023controlcom}) learns object-to-object mapping conditioned on a background image with specified placement. 
They train a diffusion model based on massive triplets of foreground images, background images with bounding boxes, and ground-truth real images, so that it can be directly applied to a new pair of foreground image and background image with bounding box to produce a composite image at test time. 
However, the generated results often fail to preserve the details of foreground object. 
Additionally, these methods only support one foreground image, lacking the ability to take advantage of multiple reference images for the same object which could provide complementary useful information. 

\begin{figure*}[t]
\centering
\includegraphics[width=0.9\linewidth]{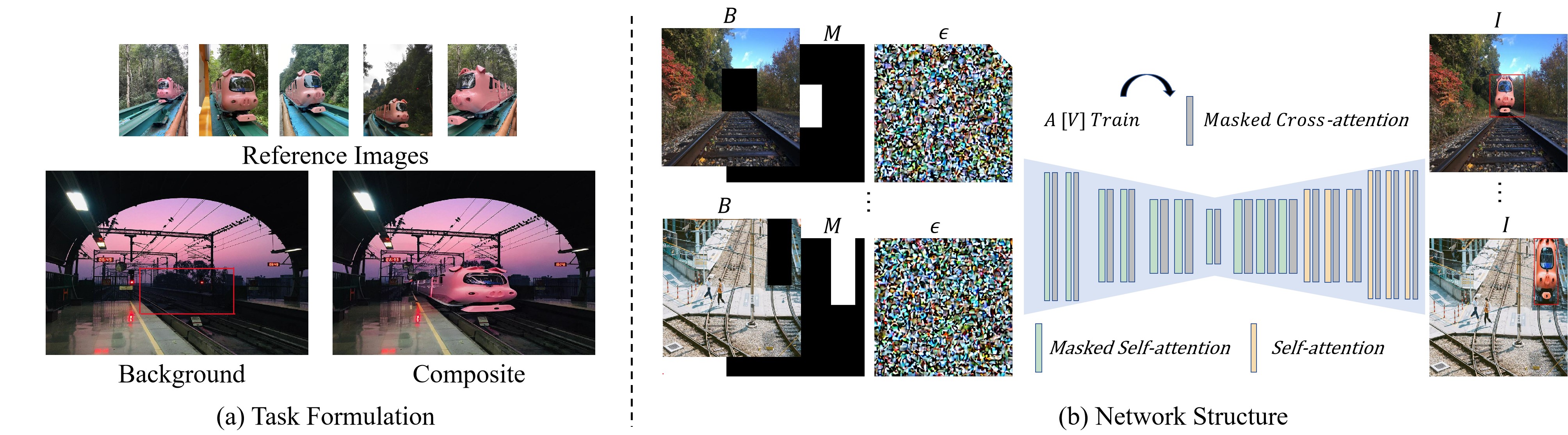}
\caption{(a) Given a background image with a bounding box and multiple reference images of the foreground object, we aim to produce a composite image, in which the foreground object is seamlessly and reasonably placed in the bounding box. (b) We finetune a text-guided inpainting model using multiple reference images of the same object, in which the text prompt contains a special token $[V]$ associated with this object. All the cross-attention layers and some selected self-attention layers are masked.}
\label{fig:task}
\end{figure*}

The second group \cite{li2023dreamedit} learns token-to-object mapping conditioned on a background image  with specified placement. 
They finetune a pretrained diffusion model on a few reference images of the same object, which enables compositing this object into other background images. 
For example, DreamEdit~\cite{li2023dreamedit} first finetunes a pretrained text-to-image generation model on a few reference images of one object to associate this object with one special token, which is the same as DreamBooth~\cite{dreambooth}. 
Then, DreamEdit segments the generated foreground and combines it with the background through the diffusion process. 
The above procedure is repeated multiple times for better visual effect. 
The pipeline in \cite{li2023dreamedit} is tedious with iterative segmentation and composition.
Moreover, the foreground generation is not conditioned on the background, so the geometric and semantic compatibility between foreground and background cannot be guaranteed. 

In this paper, we follow the second group which has better ability of preserving details and harnessing an arbitrary number of reference images. 
Similar to \cite{dreambooth,li2023dreamedit}, we finetune the model on a few reference images of one object to associate this object with one special token.
Differently, our method is based on text-guided image inpainting model instead of text-to-image generation model. 
Since we adapt \textbf{Dream}Booth~\cite{dreambooth} to image \textbf{com}position task, we name our method as DreamCom.
As shown in \Cref{fig:task} (b), we treat image composition as text-guided image inpainting task.
Given a background with a bounding box and a text prompt with a special token $[V]$, the text-guided image inpainting model aims to fill the bounding box with the foreground object associated with the special token $[V]$.
In the text-guided image inpainting model~\cite{stablediffusion}, the denoising U-Net takes the concatenation of masked background, bounding box mask, and noisy latent image as input, while the text prompt is injected into U-Net through cross-attention. 
As shown in \Cref{fig:task} (a),  given multiple reference images of the same object, 
the text-guided inpainting model is finetuned on these reference images, so that this object is associated with the special token $[V]$. 
At test time, given a new background with a bounding box, the customized model can generate this foreground object in the bounding box. 

In the text-guided inpainting model, background provides necessary guidance information (\eg., illumination, viewpoint) to ensure the compatibility between foreground and background. However, we observe that the background may negatively affect foreground generation in the following two ways. 
Firstly, the spurious correspondence between background and text prompt in the cross-attention layers could deactivate the foreground  generation. For example, when the text prompt is ``A $[V]$ horse" and the background has a horse, the text prompt may attend to the background horse and no horse is generated in the bounding box. 
To address this issue, we propose masked cross-attention, which constrains the correspondence between text prompt and image regions.
Secondly, the interaction between foreground region and background region through self-attention layers may make the background color seep into the foreground region, leading to undesired foreground color alteration. To address this issue, we propose masked self-attention, which blocks the foreground-background interaction in the first few self-attention layers. Note that the last few self-attention layers are maintained to ensure the compatibility between foreground and background.  
Compared with DreamEdit\cite{li2023dreamedit}, DreamCom does not require iterative segmentation and compositing.
Furthermore, our generated foreground is more compatible with the background and more faithful to the foreground objects in reference images.

We conduct experiments on DreamEditBench~\cite{li2023dreamedit}, which only has $220$ background images and $30$ unique foreground object from $15$ categories. 
To supplement this dataset, we construct a new dataset named as \textbf{Mu}lti-\textbf{re}ference \textbf{Com}position~(MureCom), which has $640$ background images and $96$ unique foreground object from $32$ categories. 
Experimental results on both DreamEditBench and MureCom reveal the outstanding performance of DreamCom.
The major contributions are summarized as: 1) We make the first attempt at image composition based on text-guided inpainting diffusion model. 2) We propose masked cross-attention and masked self-attention to avoid negative background interference. 3) We contribute a new image composition dataset MureCom with multiple reference images for each object.

\section{Related Work}
\subsection{Image Composition} \label{sec:image_composition}

A composite image could be easily acquired by splicing the foreground object from one image and pasting it on another background image, but the obtained composite image could be of poor quality due to the inconsistency between foreground and background. The inconsistency include appearance inconsistency, geometry inconsistency, and semantic inconsistency, as discussed in~\cite{compositionsurvey}. 

For appearance inconsistency, image blending~\cite{poissonblending,deepblending,zhang2021deep} worked on addressing the unnatural boundary so that the foreground is seamlessly blended into the background. Image harmonization~\cite{cdtnet,pctnet,dccf,dovenet,deepharmonization,xiaodong2019improving,sofiiuk2021foreground,regionaware,intriharm,guo2021image,bao2022deep,hang2022scs,jiang2021ssh,liang2021spatial,RenECCV2022,Harmonizer,WangCVPR2023,LEMaRT} studied on adjusting the foreground illumination to create a harmonious composite image. Moreover, shadow generation~\cite{shadow1,shadow2,shadow3} attempted to generate plausible shadow on the background for the inserted foreground. 
For geometry and semantic inconsistency, object placement~\cite{graconet,instanceplacement,stgan,sfgan,gccgan} aimed to predict the reasonable location, scale, and perspective of foreground given a pair of foreground and background. 

Early works on image composition usually only focus on one sub-task or performs multiple sub-tasks sequentially. Recently, generative image composition \cite{paintbyexample,objectstitch,affordanceinsertion} proposed to solve all the issues with a unified model. However, the results of \cite{paintbyexample,objectstitch} are not faithful to input foreground. DreamEdit~\cite{li2023dreamedit} has a tedious pipeline with iterative segmentation and composition.

\subsection{Subject-driven Image Generation and Editing} \label{sec:subject_image_editing}
Subject-driven image generation refers to generating images with specific subjects, while subject-driven image editing refers to replacing or inserting  specific subjects in background images. GAN-based works~\cite{interpretinggan,deepblending,hyperstyle} on subject-driven editing proposed to generate specific subjects by manipulating the latent representations in the generator. 
With the rapid development of diffusion models, text-guided subject-driven image generation~\cite{textualinversion,dreambooth,customdiffusion} has attracted increasing research interest. They proposed to leverage multiple customized images for concept learning and embed the concept into text prompts, so that the images with specified concept can be generated based on the text prompts. 
Some other works~\cite{elite,encodertuning} proposed to extract the representation of customized images using encoders. These subject-driven image generation methods cannot be directly applied to our task, because they cannot use the provided background and control the foreground placement. Additionally, some works on text-guided subject-driven image editing~\cite{photoswap,customedit,blipdiffusion} proposed to replace the objects in the background images based on pretrained diffusion model. 

It can be seen that lots of works have been done for various types of image generation and editing tasks based on diffusion model. In this work, we focus on generative image composition \cite{paintbyexample,objectstitch,affordanceinsertion,li2023dreamedit}, which targets at a realistic composite image given a background image with a bounding box and a foreground image. 

\section{Preliminary} \label{sec:prelim}

Our method is built upon Stable Diffusion (SD)~\cite{stablediffusion} model tailored for text-guided image inpainting. SD is a latent diffusion model consisting of an auto-encoder and a denoising U-Net, which are trained in two stages. 
In the first stage, an encoder \(\mathcal{E}\) projects images \(\bm{I}\) into latent space \(\bm{z}_0=\mathcal{E}(\bm{I})\) and a decoder \(\mathcal{D}\) reconstructs the original images \(\bm{\hat{I}}=\mathcal{D}(\bm{z}_0)\). The trained auto-encoder is fixed and provides latent space for the second stage.   

In the second stage, a denoising U-Net~\(\epsilon_{\theta}\)~\cite{ddpm} is trained in latent space from the first stage. In particular,  $T$-step noise is added to latent feature~\(\bm{z}_0\) to produce \(\bm{z}_t; t=1,2,...T\). Then, the denoising U-Net is trained using the following denoising loss:
\begin{equation} \label{eqn:prelim}
\!\mathcal{L}_{LDM}\!:=\!\mathbb{E}_{\epsilon\sim \mathcal{N}(0,1), t}\!\left[\!\left\|\epsilon\!-\!\epsilon_{\theta_1}\!\left(\bm{z}_{t}, t, \bm{M}, \bm{b}, \tau_{\theta_2}(y) \right)\!\right\|_{2}^{2}\!\right]\!,
\end{equation}
where $\epsilon\sim \mathcal{N}(0,1)$ is the noise added to the latent feature~\(\bm{z}_0\) in each step, $\epsilon_{\theta_1}$ is the denoising U-Net that predicts the noise~$\epsilon$ in the current step $t$. \(y\) is the text prompt and \(\tau_{\theta_2}\) is the text encoder that projects the text prompt into embeddings. $\bm{M}$ is the inpainting mask indicating the regions to be inpainted. \(\bm{b}=\mathcal{E}(\bm{B})\) is the latent feature map of masked background $\bm{B}$, which is obtained by removing the content within the mask in $\bm{I}$. The denoising U-Net \(\epsilon_{\theta_1}\) takes \(\bm{z}_t\), $t$, $\bm{M}$,  $\bm{b}$, and $y$ as input, to estimate the noise at step $t$. Specifically, \(\bm{z}_t\),  $\bm{M}$, and $\bm{b}$ are concatenated channel-wisely, leading to 9-channel input for the denoising U-Net. \(\tau_{\theta_2}(y)\) is injected into the transformer in U-Net via cross-attention. 

During inference, we sample random Gaussian noise as initial \(\bm{z}_T\).  The denoising U-Net \(\epsilon_{\theta_1}\) estimates the noise at step $T$. This procedure is iteratively executed to estimate the noise at each denoising step \(t\), so that the latent feature is gradually denoised and recovers the noise-free latent feature \(\bm{z}_0\).
Finally, \(\bm{z}_0\) is sent into the decoder \(\mathcal{D}\) to produce the image. 
For more details about the training and inference of Stable Diffusion, please refer to~\cite{stablediffusion}.

\section{Our Method}\label{sec:method}
\subsection{The Overall Network}

As illustrated in \Cref{fig:task} (b), our method is built upon text-guided image inpainting model.
Inspired by DreamBooth~\cite{dreambooth}, we first insert a special token $[V]$ into the text prompt and then associate this token with the given object by finetuning the pretrained diffusion model on a few  (\eg, 3 to 5) reference images containing the object. 
For example, given the train in \Cref{fig:task} (b), the text prompt $y$ is set as ``A $[V]$ train". 
In each reference image, we extract the bounding box of foreground train to get the inpainting mask $\bm{M}$, followed by erasing the content within bounding box to get the masked background $\bm{B}$. 
After that, we finetune the pretrained inpainting model using the loss function in \Cref{eqn:prelim} with the masked background, inpainting mask, and noisy latent feature being the input. 
To enrich the limited training set (\eg, 3 to 5 reference), we perform data augmentation including random cropping and scaling.

After finetuning the inpainting model, given a new background image with a bounding box, we can get $\bm{M}$ and $\bm{B}$. By using text prompt $y$ containing the special token $[V]$, we start from random Gaussian noise $\bm{z}_T\sim \mathcal{N}(0,1)$ and produce the composite image through the denoising process.

\begin{figure}[t]
\centering
\includegraphics[width=0.90\linewidth]{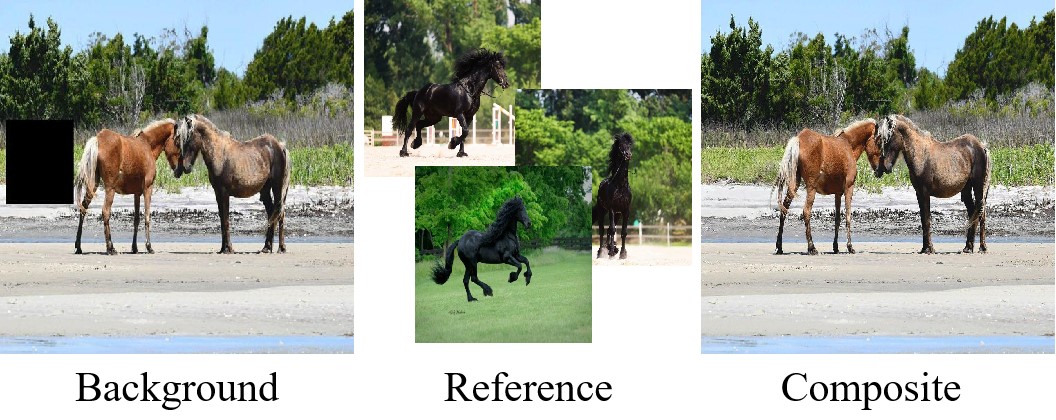}
\caption{An example of spurious correspondence between background and text prompt. From left to right, we present the masked background image, reference images, and the generated image with the text prompt ``A $[V]$ horse''. }
\label{fig:cross_attention_mask}
\end{figure}

\begin{figure}[t]
\centering
\includegraphics[width=0.90\linewidth]{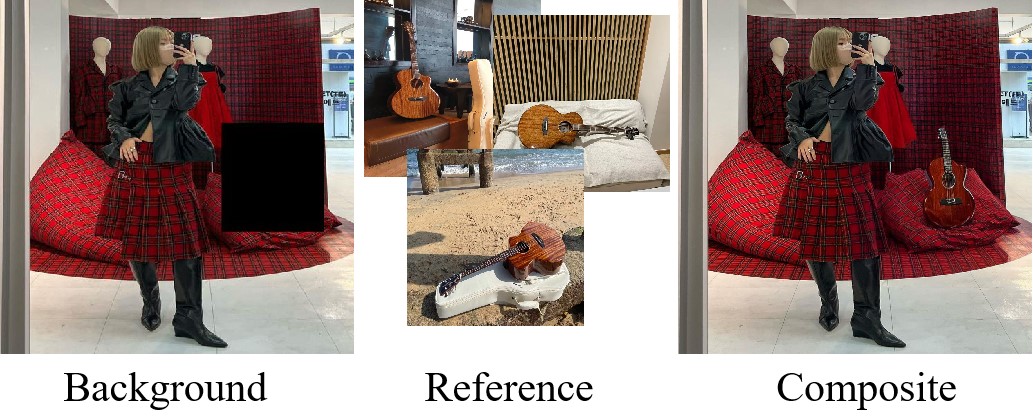}
\caption{An example of foreground color disrupted by the background image. From left to right, we present the masked background image, reference images, and the generated image with the text prompt "A $[V]$ guitar".}
\label{fig:self_attention_mask}
\end{figure}

\subsection{Masked Cross-attention}\label{sec:method:maskca}

During both training and inference, we observe that the pretrained inpainting model\footnote{https://huggingface.co/runwayml/stable-diffusion-inpainting} is trained with global text prompts, \ie, the prompts describe the entire image rather than just the inpainted region. 
Consequently, if we retain the original interaction mechanism between intermediate features from U-Net and the text prompt in the pretrained inpainting model, there could be potential interference from the background elements affecting the foreground. 
For example, as illustrated in \Cref{fig:cross_attention_mask}, if the background contains a horse, the text prompt ``A photo of a $[V]$ horse'' may inadvertently focus on the horses in the background, resulting in the failure to generate a horse within the designated bounding box. 
Therefore, to ensure that the text prompt exclusively influences the foreground generation while ignoring the background, we incorporate the foreground mask (inpainting mask) into the text-image cross-attention mechanism.

Specifically, the original attention map \textbf{A}$^{ca} \in \mathbb{R}^{l_q \times l_k}$ is calculated through ${\rm softmax} \left( \frac{\mathbf{Q} \times \mathbf{K}^{T}}{\sqrt{d}} \right)$, where \textbf{Q} represents the intermediate feature from U-Net, \textbf{K} is the text prompt embedding $\tau_{\theta_2}(y)$. $l_q, l_k$, and $d$ are the length of \textbf{Q}, the length of \textbf{K} and the feature dimension of \textbf{Q} and \textbf{K}, respectively.
The cross-attention mask $\bm{M}_{\mathbf{A}}^{ca}$ have the same shape as \textbf{A}$^{ca}$, where the positions corresponding to $\tau_{\theta_2}(y)$ and the foreground region $\mathbf{Q}_{fg}$ (\emph{resp.}, background region $\mathbf{Q}_{bg}$) are set as 1 (\emph{resp.}, 0). 
Therefore, the masked cross-attention map $\mathbf{A}_{fg}^{ca}$ is formulated as
\begin{equation}
\mathbf{A}_{fg}^{ca} = \bm{M}_{\mathbf{A}}^{ca} \circ \mathbf{A}^{ca}, \label{equ:maskedsa}
\end{equation}
in which $\circ$ means element-wise multiplication.

\subsection{Masked Self-attention}

Another notable observation is the tendency for the colors of foreground objects to be influenced by the background image. 
This phenomenon is likely due to the interaction between foreground and background features via self-attention layers. 
For example, as depicted in \Cref{fig:self_attention_mask}, if the background image is dark red, the color of the generated foreground object may shift from its original color to dark red. 
A potential solution to mitigate the foreground-background interference is to employ attention mask in the self-attention layers, akin to the masked cross-attention  in \Cref{sec:method:maskca}. 
However, this approach might compromise the compatibility between the foreground and background, as it limits their interaction.
In order to accomplish the foreground-background compatibility while achieving fine control of the foreground color, we investigate the impact of self-attention in different U-Net layers.

Based on experimental observations, the outer layers of the U-Net decoder dominate the overall harmonization, without which the generated foreground would be inharmonious with the background. 
Furthermore, all layers of the U-Net influence the color of the foreground object. 
Based on these insights, we decide not to apply the self-attention mask in the last $6$ layers of the U-Net decoder. 
For the remaining layers, adjustments can be made during inference, as illustrated in \Cref{fig:task} (b). 
Specifically, we define two types of features: the foreground feature $\mathbf{Q}_{fg}$ and the background feature $\mathbf{Q}_{bg}$, corresponding to the foreground and background regions, respectively. 
The value in the self-attention mask is set to 1 (\emph{resp.}, 0) if the position correlates with $\mathbf{Q}_{fg}$-$\mathbf{Q}_{fg}$ or $\mathbf{Q}_{bg}$-$\mathbf{Q}_{bg}$ (\emph{resp.}, $\mathbf{Q}_{fg}$-$\mathbf{Q}_{bg}$ and $\mathbf{Q}_{bg}$-$\mathbf{Q}_{fg}$). 
The computation of masked self-attention map is similar to that of masked cross-attention map, as detailed in \Cref{equ:maskedsa}. 
Further details regarding the impact of the self-attention mask are in \Cref{sec:sa_impact} and Supplementary Material.

\section{Dataset Construction}
In DreamEdit~\cite{li2023dreamedit}, they provide the DreamEditBench dataset, offering a total of 440 background images, out of which 220 background images are suitable for our task. 
For the 220 background images, there are a total of 22 background sets, each consisting of 10 images.
The reference images corresponding to the backgrounds in this dataset are sourced from the Dreambooth dataset~\cite{dreambooth}. 
There are 30 foreground objects distributed in 15 categories, in which each foreground object has a 3 to 5 reference images. 
Among them, some foreground objects share a common background set, while others have their own unique background sets. 
Totally, there are $300=30\times 10$ foreground-background pairs.

\begin{figure}[t]
\centering
\includegraphics[width=0.99\linewidth]{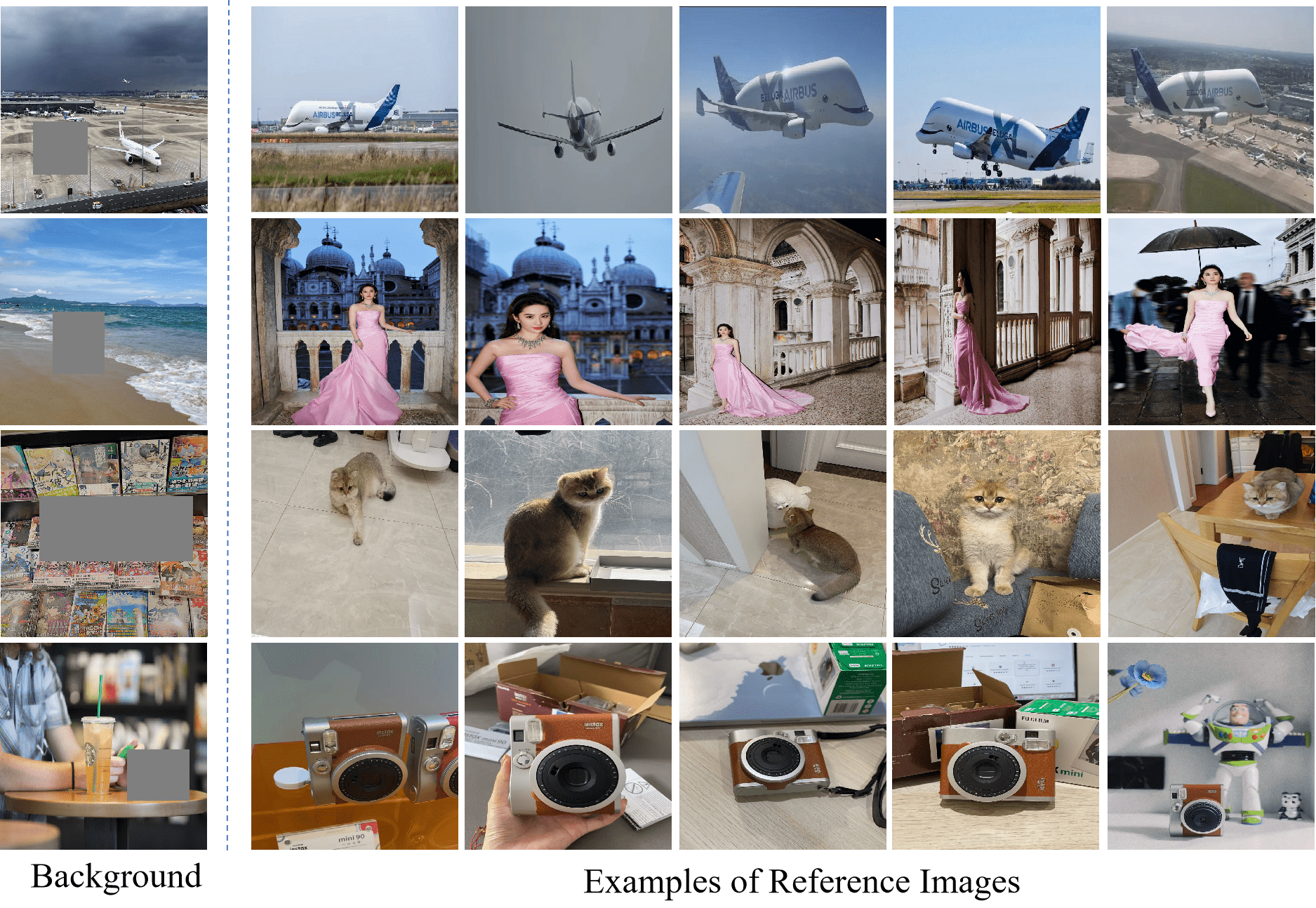}
\caption{Several examples from our MureCom dataset. In each example, we show a background image with a bounding box, and five reference images of one foreground object.}
\label{fig:dataset}
\end{figure}

To enrich the diversity of DreamEditBench dataset, we construct a new dataset, \textbf{Mu}lti-\textbf{re}ference \textbf{Co}mposition (MureCom). 
The MureCom dataset is constructed with the following steps:
1) we collect $640$ background images from R-FOSD dataset \cite{FOS}, which is associated with $32$ different foreground categories and each foreground category contains $20$ background images;
2) for each background image, we manually annotate a bounding box suitable for placing the object from the corresponding foreground category;
3) for each foreground category, we collect $3$ unique foreground objects with $5$ reference images for each foreground object. 
Since each foreground category has $20$ background images and $3$ unique foreground objects, we can have $1,920 = 32\times 20 \times 3$ foreground-background pairs in total. 
In \Cref{fig:dataset}, we provide some examples of background images with bounding boxes and foreground objects with reference images.   

\section{Experiments}

\subsection{Datasets and Implementation Details}
We conduct experiments on  DreamEditBench and MureCom datasets. 
On both datasets, we first finetune the pretrained diffusion model for each foreground object by using the corresponding reference images to associate the special token $[V]$ in the text prompt with the foreground object. Then, we apply the finetuned model to the background images with the text prompt containing $[V]$ to generate the composite image.
For data augmentation, when scaling down, the outer rings are filled with white pixel (\ie, [0, 0, 0] for RGB channels) to maintain the same resolution. 
When scaling up or randomly cropping, the foreground object occupies no less than $80\%$ of the original size.
We use pretrained diffusion model Stable-Diffusion-Inpainting provided by RunwayML. 
Its U-Net has 5 additional input channels and is trained on image inpainting task. 
For self-attention, we apply masked self-attention except the last 6 layers of the U-Net decoder. 
We implement DreamCom and baselines with PyTorch 1.11.0 on Ubuntu 20.04 LTS equipped with one GeForce RTX 3090 GPU, Intel Xeon Silver 4116 CPU, and 32GB of memory.

\begin{table}
\centering
\resizebox{\columnwidth}{!}{%
\setlength\tabcolsep{10pt}
\begin{tabular}{l|cc|cc}
\hline
\multirow{2}*{Method} & \multicolumn{2}{c|}{Foreground} & \multicolumn{2}{c}{Background} \\
  &$\mathrm{DINO}$↑ &$\mathrm{CLIP}$↑  &$\mathrm{LPIPS}$↓   &$\mathrm{SSIM}$↑   \\ 
\hline \hline
ObjectStitch & 0.599 & 0.801   &0.141 & 0.764  \\
PbE  & 0.514 & 0.768  &0.149 & 0.755   \\
DreamEdit  & 0.490 & 0.733  &0.207 & 0.740  \\
\hline
DreamCom  & 0.669 & 0.805  &0.154 & 0.776   \\
\hline
\end{tabular}
}
\caption{Quantitative comparison on DreamEditBench dataset.}
\label{tab:baseline1}
\end{table}

\begin{table}
\centering
\resizebox{\columnwidth}{!}{%
\setlength\tabcolsep{10pt}
\begin{tabular}{l|cc|cc}
\hline
\multirow{2}*{Method} & \multicolumn{2}{c|}{Foreground} & \multicolumn{2}{c}{Background} \\
 &$\mathrm{DINO}$↑ &$\mathrm{CLIP}$↑  &$\mathrm{LPIPS}$↓   &$\mathrm{SSIM}$↑  \\ 
\hline \hline 
ObjectStitch & 0.502 & 0.762  &0.150   &0.736  \\
PbE  & 0.414 & 0.733  &0.172   &0.708    \\
DreamEdit & 0.301 & 0.657   &0.232   &0.692  \\
\hline
DreamCom  & 0.584 & 0.778  &0.174  &0.731  \\
\hline
\end{tabular}
}
\caption{Quantitative comparison on our MureCom dataset.}
\label{tab:baseline2}
\end{table}

\subsection{Evaluation Metrics}
Following Dreambooth~\cite{dreambooth} and DreamEdit~\cite{li2023dreamedit}, we use DINO~\cite{caron2021emerging} and CLIP-I~\cite{radford2021learning} scores to evaluate the fidelity of foreground. 
Specifically, we generate five composite images for each foreground-background pair, followed by calculating the DINO and CLIP-I scores based on all foreground-background pairs.
Specifically, the foreground of each generated image is evaluated against the foregrounds of all reference images to obtain the scores, which are averaged as the final score of each foreground-background pair. 
Besides, we further evaluate the background preservation using SSIM~\cite{ssim} and LPIPS~\cite{lpips} scores.

Following~\cite{paintbyexample}, we further conduct user study with average score to measure the  compatibility of composite image and the fidelity of foreground. 
In detail, DreamEditBench and MureCom datasets consist of 30 and 32 foreground categories, respectively. 
For each foreground category, we randomly select 2 backgrounds to construct 60 and 64 foreground-background pairs.
Each background-foreground pair is then used to generate four composite images by PbE, ObjectStitch, DreamEdit, and our DreamCom.
Therefore, we collect 124=60+64 groups of images, with each group containing a background image with a bounding box, a foreground reference image, and four composite images.
Then we hire 50 annotators to independently rate the compatibility and fidelity of composite images on the scale [1,2,3,4] (from best to worst) within each group. 
Note that when rating the compatibility, these annotators are asked to judge from pose, boundary, and illumination.

\begin{figure}[t]
\begin{minipage}{\linewidth}
\centerline{\includegraphics[width=1\textwidth]{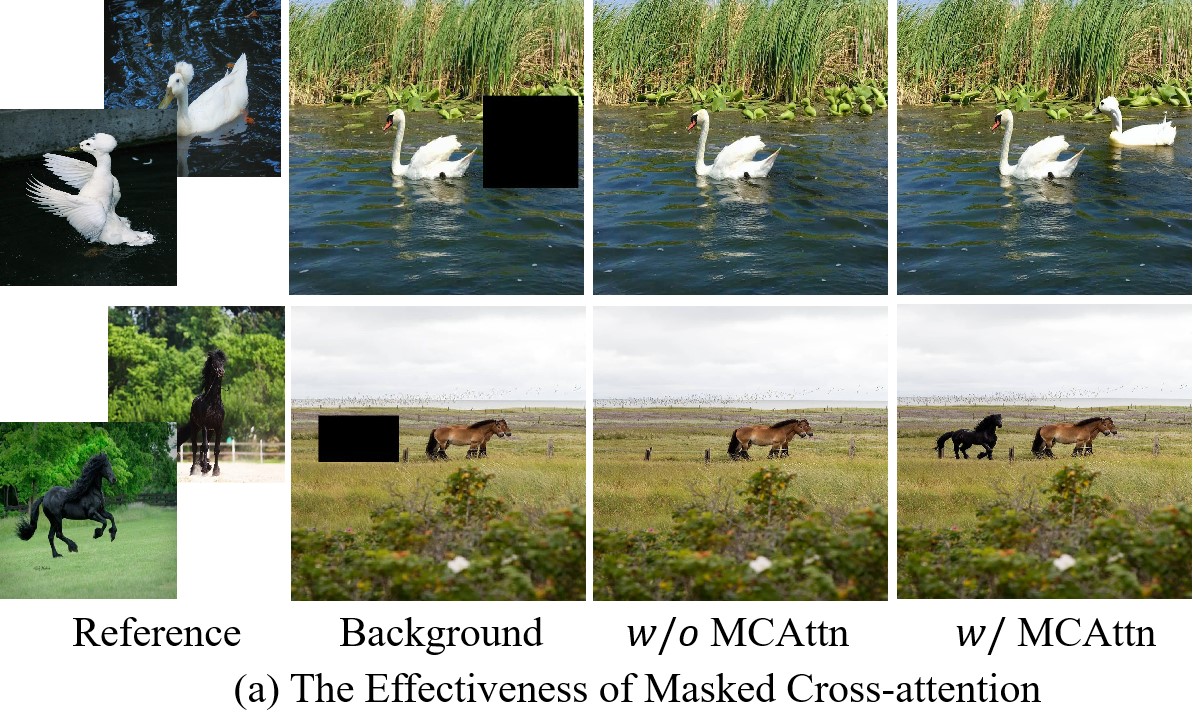}}
\end{minipage}

\hspace{0.1mm}
\begin{minipage}{\linewidth}
\centerline{\includegraphics[width=1\textwidth]{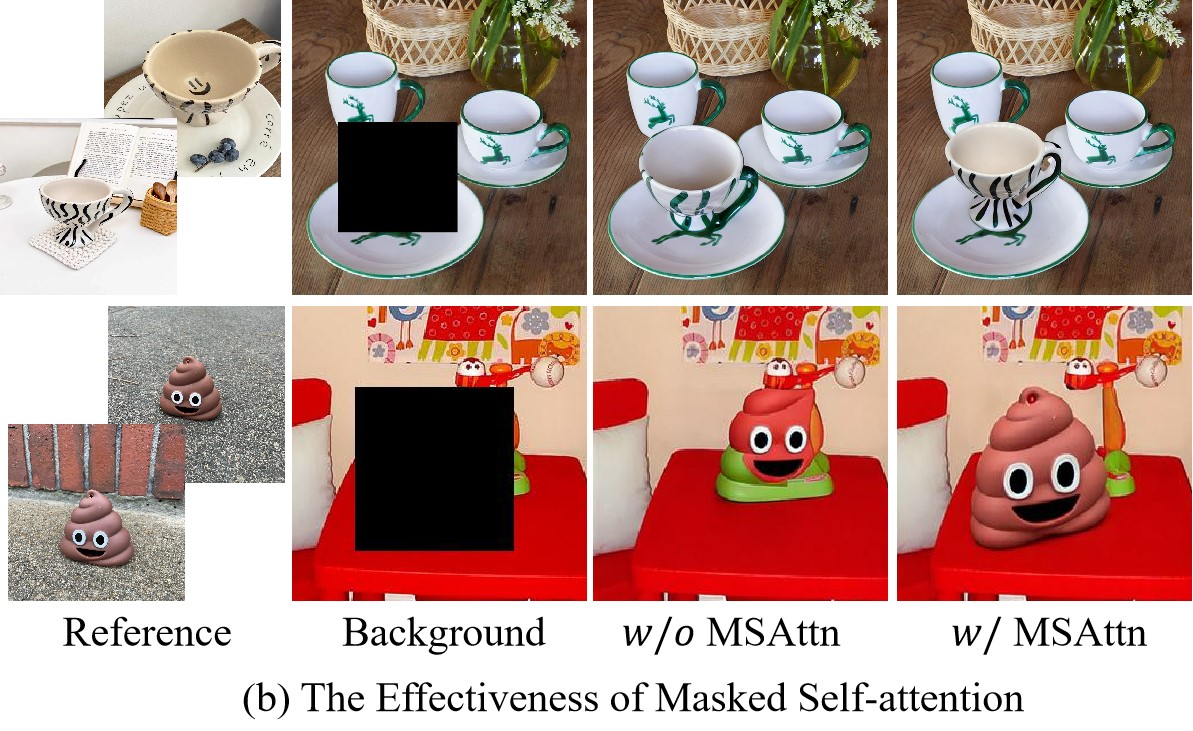}}
\end{minipage}
\caption{The subfigure (a) demonstrates the effectiveness of masked self-attention~(MCAttn).
The subfigure (b) demonstrates the effectiveness of masked self-attention~(MSAttn) in maintaining original color while preserving overall harmony.}
\label{fig:exp_attn}
\end{figure}

\begin{figure*}[t]
\centering
\includegraphics[width=0.88\linewidth]{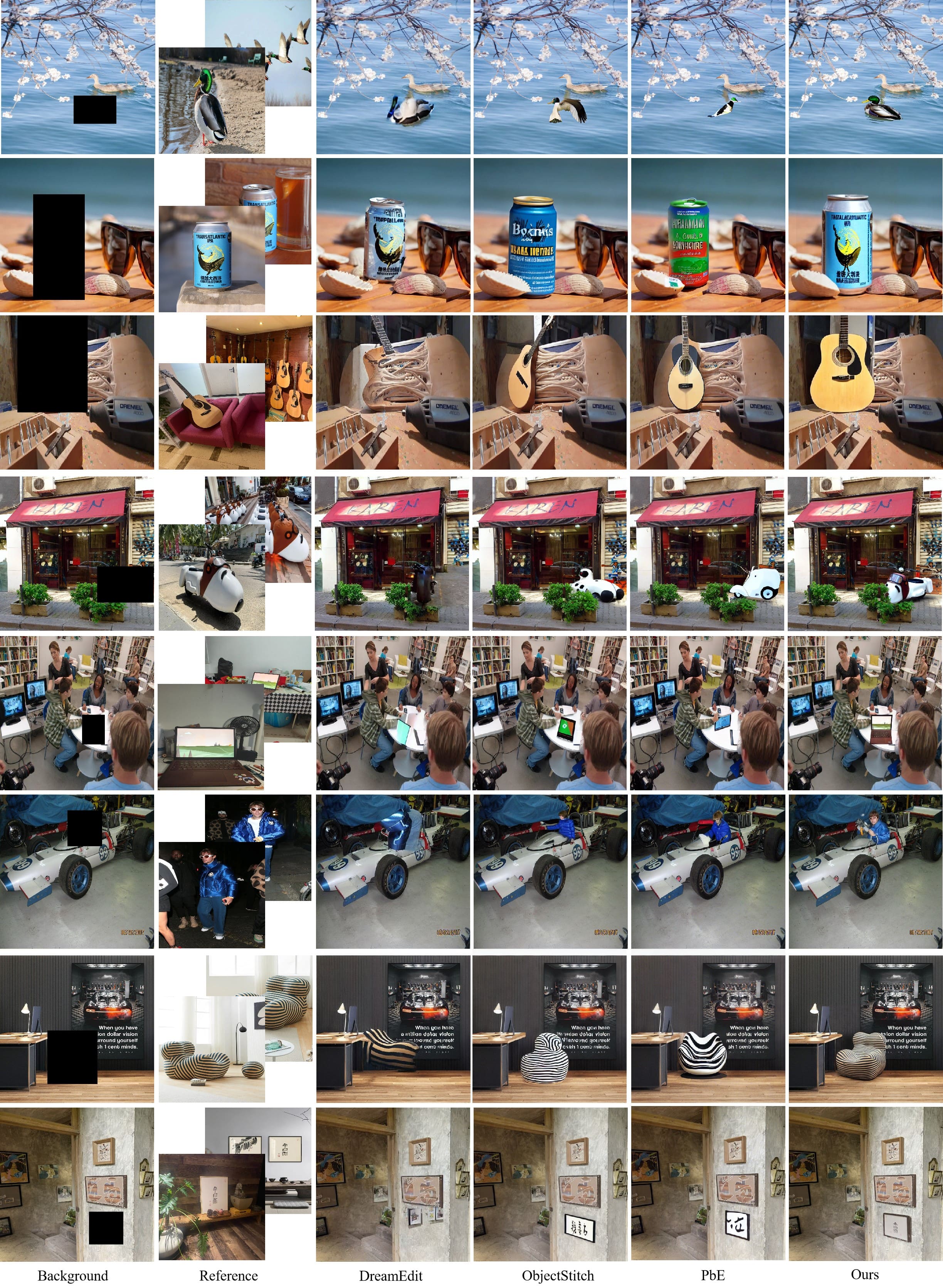}
\caption{Visualization of different methods. In each row, we show the background image with bounding box, two reference images of foreground object, and the results of DreamEdit \cite{li2023dreamedit}, ObjectStitch \cite{objectstitch}, PbE \cite{paintbyexample}, and our DreamCom. }
\label{fig:visual_comparison}
\end{figure*}

\subsection{Comparison with Baselines}

We compare our DreamCom with PbE~\cite{paintbyexample}, ObjectStitch~\cite{objectstitch}, and DreamEdit~\cite{li2023dreamedit}. 
For DreamEdit, we follow the same procedures in DreamEdit by first finetuning the pretrained stable diffusion model for each foreground object and then utilizing GLIGEN to initialize the background image during inference. 
After that, the finetuned DreamEdit is employed to fill in the bounding box of the background. 
The generated result is segmented and a new mask is extracted as the input of next round, where the generation process is repeated for five times. 
For PbE, we directly use the released model pretrained on Open Images dataset~\cite{openimages} for evaluation. 
For ObjectStitch, we re-implement this model, and train it on Open Images dataset~\cite{openimages} for evaluation.

The experimental results on DreamEditBench and MureCom are reported in \Cref{tab:baseline1} and \Cref{tab:baseline2}, respectively. 
For DINO and CLIP scores of foreground objects, our DreamCom achieves the highest scores, followed by ObjectStitch, PbE, and Dreamedit, respectively, which suggests that the foreground objects generated by our DreamCom is capable of preserving more details in the original foreground objects than the other methods. Compared with CLIP-I, DINO aims to distinguish different foreground objects with similar descriptions, capturing more details of foreground object. Therefore, the advantage of our method is more notable when using DINO score. 
Besides, we also find that ObjectStitch and PbE can only preserve the same semantic category instead of object details. Meanwhile, DreamEdit performs poorly and struggles to generate ideal results for many challenging backgrounds. 

For SSIM and LPIPS, our DreamCom also achieves comparable scores with PBE and ObjectStitch, which indicates that our DreamCom can also effectively preserve enough details in the background image. 
In contrast, due to mask dilation, DreamEdit fails to retain the background, leading to significant variations in the background.

\begin{table}
\centering
\resizebox{\columnwidth}{!}{%
\setlength\tabcolsep{5pt}
\begin{tabular}{l|cc|cc}
\hline
\multirow{2}*{Method} & \multicolumn{2}{c|}{DreamEditBench} & \multicolumn{2}{c}{MureCom} \\
  &$\mathrm{Compatibility}$↓ &$\mathrm{Fidelity}$↓ &$\mathrm{Compatibility}$↓   &$\mathrm{Fidelity}$↓   \\ 
\hline \hline
ObjectStitch & 2.10 & 2.87   &2.04 & 2.32  \\
PbE           & 2.73 &3.30  &2.61 & 3.01   \\
DreamEdit  & 3.47 & 2.47  &3.37 & 3.04  \\
\hline
DreamCom  & 1.70 & 1.37  &1.98 &1.63  \\
\hline
\end{tabular}
}
\caption{User study on DreamEditBench and MureCom dataset. We calculate the average ranking of compatibility of composite images and fidelity of foregrounds. 1 is the best and 4 is the worst.}
\label{tab:user_study}
\end{table}

In the user study, the average rankings are presented in \Cref{tab:user_study}. 
Our DreamCom secures the highest average ranking in both compatibility and fidelity.
In comparison, methods like PbE and ObjectStitch are ranked similarly, but significantly lower than DreamCom. 
DreamEdit is ranked the lowest, mainly because of its sensitivity to different initialization methods and unstable iterative procedure.

In \Cref{fig:visual_comparison}, we compare the generated composite images from DreamEdit, ObjectStitch, PbE, and our DreamCom. 
We observe that the generated composite images from PbE and ObjectStitch only preserve the semantic category of the foreground object in reference image, while ignoring the appearance and textual details. 
Besides, they are more susceptible to the pose of single reference image~(row 1, with the reference image depicting a soaring goose), and the shapes of the generated objects may be distorted~(row 3, 7).
For DreamEdit, it performs reasonably well in relatively simpler backgrounds~(row 2). However, in the case of more challenging backgrounds, DreamEdit struggles to generate desired composite images~(row 3, 6, 8). 
And it is highly vulnerable to the quality of object mask extraction during iterations, which consequently leads to deteriorating results(row 5, 7).
Our approach not only preserves the appearance and textual details of the foreground objects~(especially row 2, 4, 5, 8), but also exhibits better adaptability~(pose and boundary) to more challenging backgrounds~(row 4, 6, 7, 8). 
Therefore, DreamCom can better balance foreground preservation and foreground-background compatibility,  generating more realistic composite images.

\begin{figure}[t]
\centering
\includegraphics[width=0.75\linewidth]{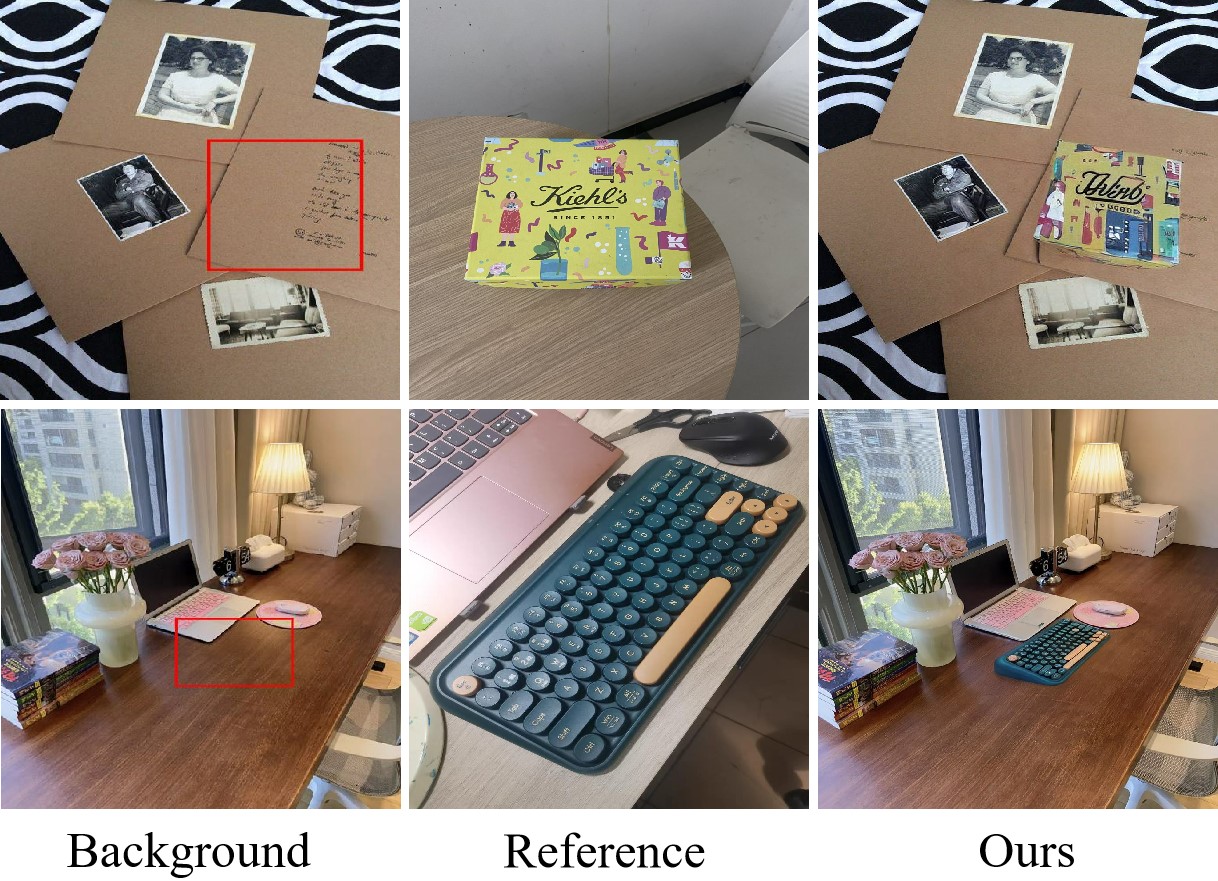}
\caption{The failure cases of our DreamCom.}
\label{fig:limitation}
\end{figure}

\subsection{Ablation Study}
In this section, we study the effect of masked cross-attention and masked self-attention on MureCom dataset. More experiments about the effect of self-attention layers and the number of reference images are in Supplementary Material.

\subsubsection{Masked Cross-attention}

To investigate the impact of masked cross-attention, we compare with using vanilla cross-attention and report the results in \Cref{tab:ablation_study} (row 1 \emph{v.s.} row 2) and in \Cref{fig:exp_attn}~(a). 
According to \Cref{tab:ablation_study}, the masked cross-attention achieves an improvement of $1.8\%$ in the DINO score and an improvement of $0.7\%$ in the CLIP score, underscoring its efficacy in preserving foreground details. 
Furthermore, as shown in \Cref{fig:exp_attn}~(a), when the background contains a object mentioned in the text prompt (row 1) or the bounding box is relatively small (row 2), the approach using vanilla cross-attention fails to generate foreground object in the bounding box. 
Conversely, equipped with masked cross-attention, the foreground objects are generated in composite image, demonstrating the effectiveness of masked cross-attention.

\begin{table}
\centering
\resizebox{0.6\columnwidth}{!}{%
\begin{tabular}{l|l|ll}
\hline
\multirow{2}*{} &\multirow{2}*{Method} & \multicolumn{2}{c}{Foreground} \\
 & &$\mathrm{DINO_{fg}}$↑ &$\mathrm{CLIP_{fg}}$↑  \\ 
\hline 
1 &DreamCom  & 0.584 & 0.778 \\
2 &-MCAttn   & 0.566  & 0.771   \\
3 &-MSAttn  & 0.573 & 0.775\\
4 &-Both   & 0.546 & 0.764\\
\hline
\end{tabular}
}
\caption{Quantitative evaluation of the impact of masked cross-attention (MCAttn) and masked self-attention (MSAttn) on our MureCom dataset.}
\label{tab:ablation_study}
\end{table}

\subsubsection{Masked Self-attention}\label{sec:sa_impact}

To investigate the impact of masked self-attention, we replace masked self-attention with vanilla self-attention and report the results in \Cref{tab:ablation_study} (row 1 \emph{v.s.} row 3) and in \Cref{fig:exp_attn}~(b). 
According to \Cref{tab:ablation_study}, masked self-attention leads to an improvement of $1.1\%$  in the DINO score and an improvement of $0.3\%$ in the CLIP score. 
This improvement is smaller than that brought by masked cross-attention. 
Considering that the main benefit of masked cross-attention is preventing the loss of foreground objects, and the main benefit of masked self-attention is preserving foreground colors, the smaller performance improvement of masked self-attention is reasonable. 
Moreover, as shown in \Cref{fig:exp_attn}~(b), the absence of masked self-attention could make the foreground object in the composite image affected by the background color or style, leading to a loss of original foreground color. 
However, with masked self-attention, the preservation of foreground color is notably enhanced. 
Additionally, we evaluate DreamCom by removing both masked self-attention and masked cross-attention, leading to the results in \Cref{tab:ablation_study} (row 1 \emph{v.s.} row 4). 
Here, DreamCom achieves the improvements of $3.8\%$ and $1.4\%$ in the DINO and CLIP scores, respectively. 

\subsection{Limitation}

We provide some failure cases of DreamCom in \Cref{fig:limitation}. 
For foreground objects with delicate details (\eg, texts on the box and patterns on the keyboard), DreamCom may fail to reconstruct those details, where the baseline methods also fail to preserve all details.
Thus, preserving delicate details is still a tough task for generative image composition.

\section{Conclusion}

In this paper, we have proposed a simple baseline for image composition, \ie, finetuning the pretrained text-guided inpainting model equipped with masked cross-attention and self-attention on a few reference images of the same foreground object. 
We have also contributed a new dataset named MureCom, which could significantly facilitate the research on multi-reference image composition.

{\small
\bibliographystyle{ieee_fullname}
\bibliography{main}
}

\end{document}


\title{Supplementary for DreamCom: Finetuning Text-guided Inpainting Model for Image Composition}

\author{Lingxiao Lu, Jiangtong Li, Bo Zhang, Li Niu\thanks{Corresponding author.} \\
Department of Computer Science and Engineering, MoE Key Lab of Artificial Intelligence, \\
Shanghai Jiao Tong University\\
{\tt \small \{lulingxiao,keep\_moving-Lee,bo-zhang,ustcnewly\}@sjtu.edu.cn}
}

\maketitle

In this document, we provide supplementary materials to support the main text. 
In \Cref{sec:roles}, we validate the impact of masked self-attention in different U-Net layers, and summarize their roles in preserving foreground information and achieving the overall compatibility. 
In \Cref{sec:number}, we conduct experiments by finetuning the model with different numbers (1-5) of reference images to verify the effect of reference image quantity.
Finally, in \Cref{sec:baseline}, we present additional visual comparison with the baselines, demonstrating the superiority of our approach.

\section{Roles of Self-attention Layers}
\label{sec:roles}
\begin{figure*}[t]
\centering
\includegraphics[width=\linewidth]{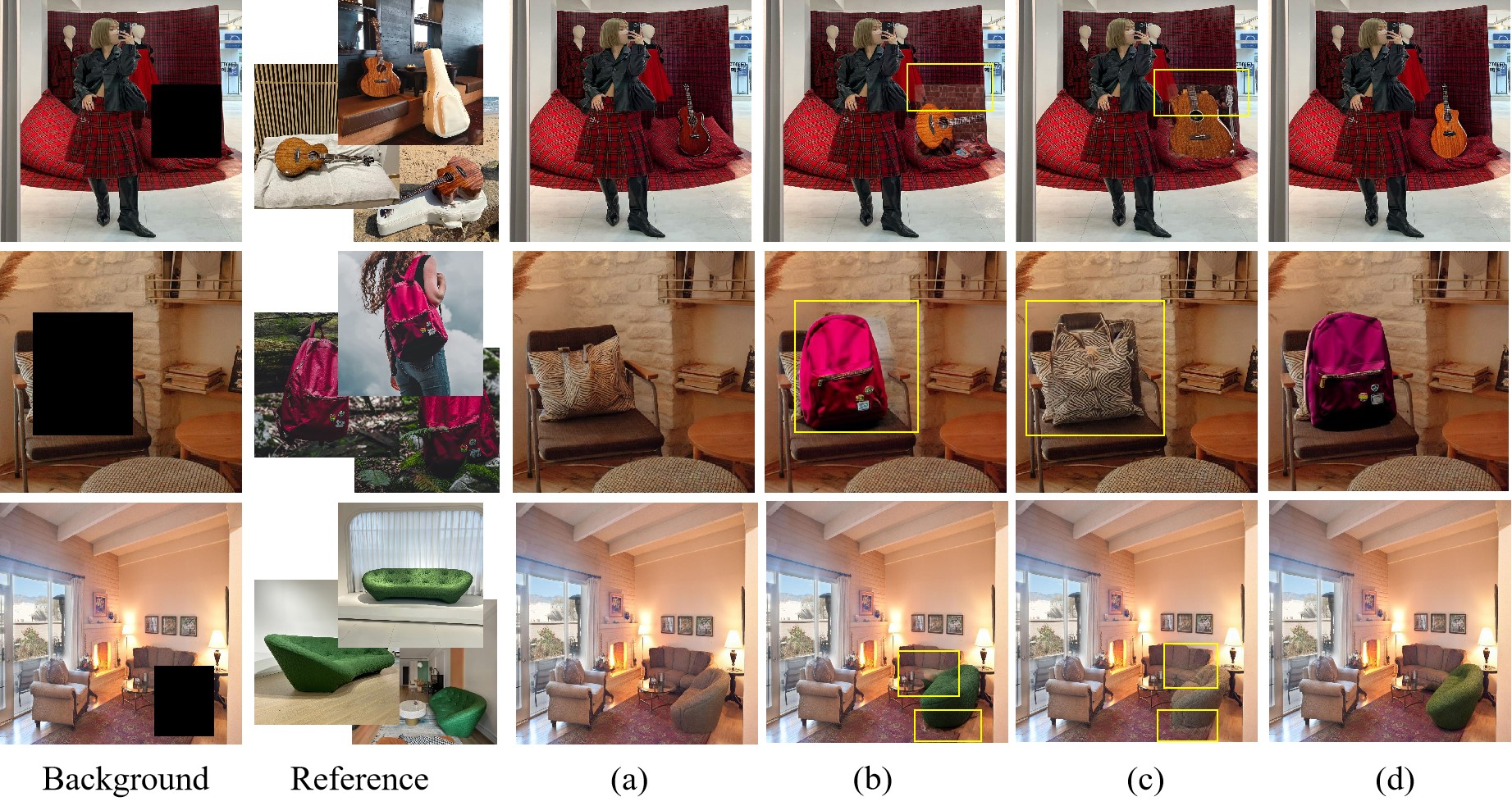}
\caption{The results of masking different layers of self-attention in U-Net. From left to right in each row, we present the background, several reference images, the result with all layers unmasked (a), the result with all layers masked (b), the result with decoder's outer layers masked and the other layers unmasked (c), the result with decoder's outer layers unmasked and the other layers masked (d).}
\label{fig:sup_self}
\end{figure*}

Here we conduct experiments to verify that the outer self-attention layers of the decoder determine the overall compatibility and masking the other layers can help solve color leaking from background to foreground. We construct the following control groups: a) all layers are unmasked, b) all layers are masked, c) decoder's outer layers are masked while the other layers are unmasked, d) decoder's outer layers are unmasked while the other layers are masked. 

As illustrated in \Cref{fig:sup_self}, for a), unmasking all layers results in the background color leaking into the foreground, causing the loss of its own color information. 
For b), masking all layers leads to notable discrepancy between the foreground and background (marked by yellow bounding box).
For c), masking the decoder's outer layers while unmasking the other layers results in the worse case. There exists notable discrepancy between the foreground and background (marked by yellow bounding box), and the background color is leaking into the foreground.
For d), unmasking the decoder's outer layers and masking the other layers could achieve the following two effects: 1) The foreground is seamlessly blended into the background, with compatible lighting and pose. 
2) The foreground is protected from background color leakage, and thus maintains its own appearance and texture details.

\section{The Number of Reference Images}
\label{sec:number}
\begin{figure*}[t]
\centering
\includegraphics[width=\linewidth]{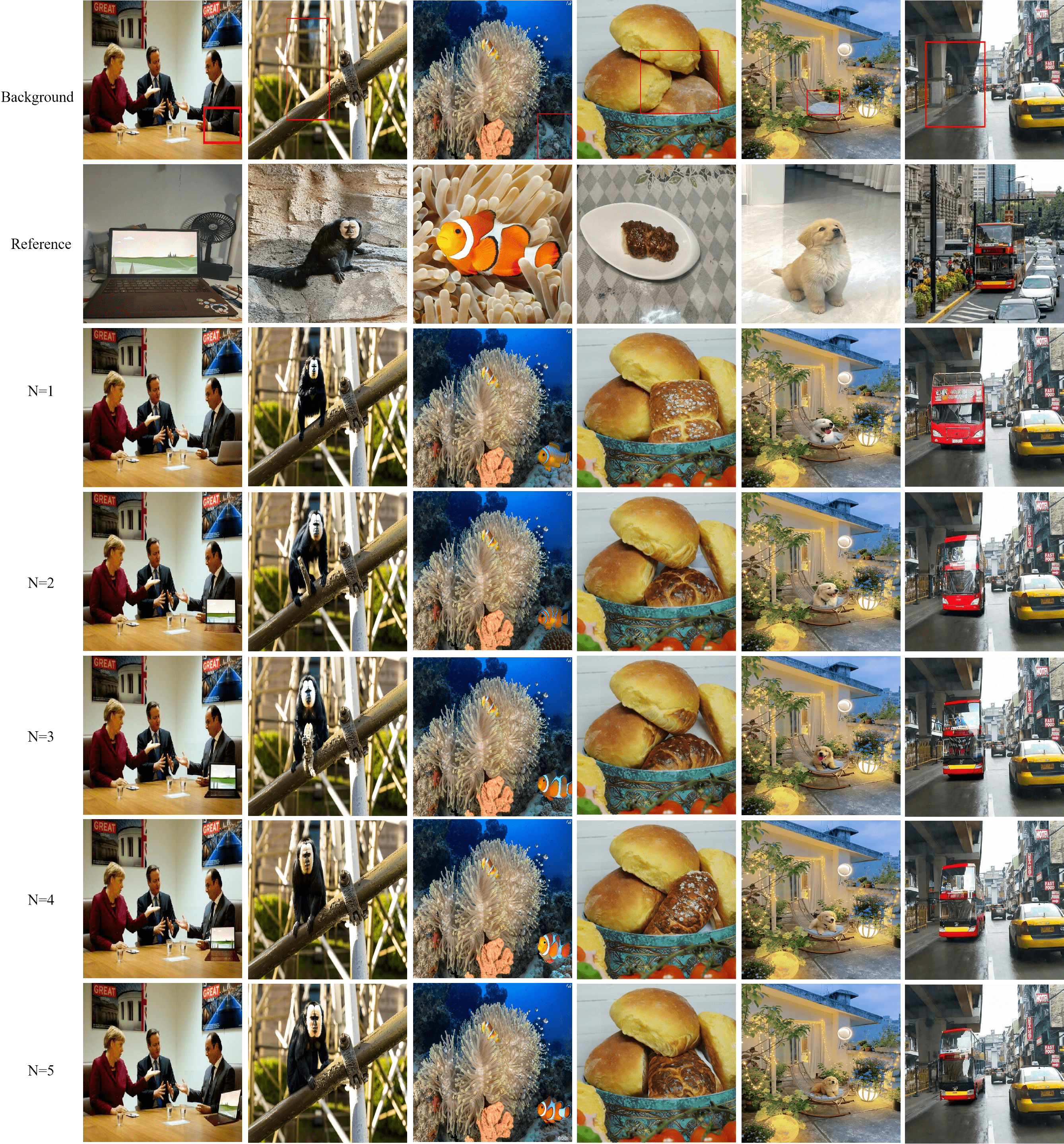}
\caption{Visualization results using different numbers of reference images per object. In each column, we show the background image with a bounding box, one reference image of foreground object, and our results.}
\label{fig:number_of_reference_images}
\end{figure*}

Recall that we use 5 reference images for each object by default. 
In this section, we investigate the impact of the number of reference images. We conduct experiments using various numbers of reference images, ranging from 1 to 5. 
As indicated by \cref{tab:number_of_reference_images}, the Clip score and Dino score usually increase as the number of reference images increases. The performances using 4 or 5 reference images are comparable. This could be attributed to the fact that for some simple backgrounds, the provided 4 reference images are already sufficient to cover diverse viewpoints and poses. However, for other backgrounds, even with 5 reference images, it may not be enough to learn appropriate viewpoints and poses that adapt well to those backgrounds, indicating the need of additional reference images. 
As shown in \cref{fig:number_of_reference_images}, the results are generally getting better as the number of reference images increases. Firstly, incorporating more references enables the model to learn from a broader range of perspectives, resulting in a more comprehensive understanding of details (column 2, 6). Secondly, by leveraging an increased number of references, the model gains access to additional information on how the object interacts with and adapts to the background. Consequently, during the generation process, it exhibits improved performance even in more challenging locations (column 1, 3, 4, 5).

\begin{table}
\centering
\begin{tabular}{>{\centering\arraybackslash}p{2cm}|>{\centering\arraybackslash}p{2cm}p{2cm}}
\hline
\multirow{2}*{Number} & \multicolumn{2}{c}{Foreground} \\
  &$\mathrm{DINO_{fg}}$↑ &$\mathrm{CLIP_{fg}}$↑   \\ 
\hline \hline
1  & 0.539  &  0.762 \\
2  & 0.573 & 0.773    \\
3 & 0.581 &  0.775 \\
4 & 0.586 &  0.777 \\
5 & 0.584  &  0.778\\

\hline
\end{tabular}
\caption{Quantitative evaluation using different numbers of reference images per object on our MureCom dataset.}
\label{tab:number_of_reference_images}
\end{table}

\section{More Comparison with Baselines}\label{sec:comp_base}
\label{sec:baseline}
\begin{figure*}[t]
\centering
\includegraphics[width=0.89\linewidth]{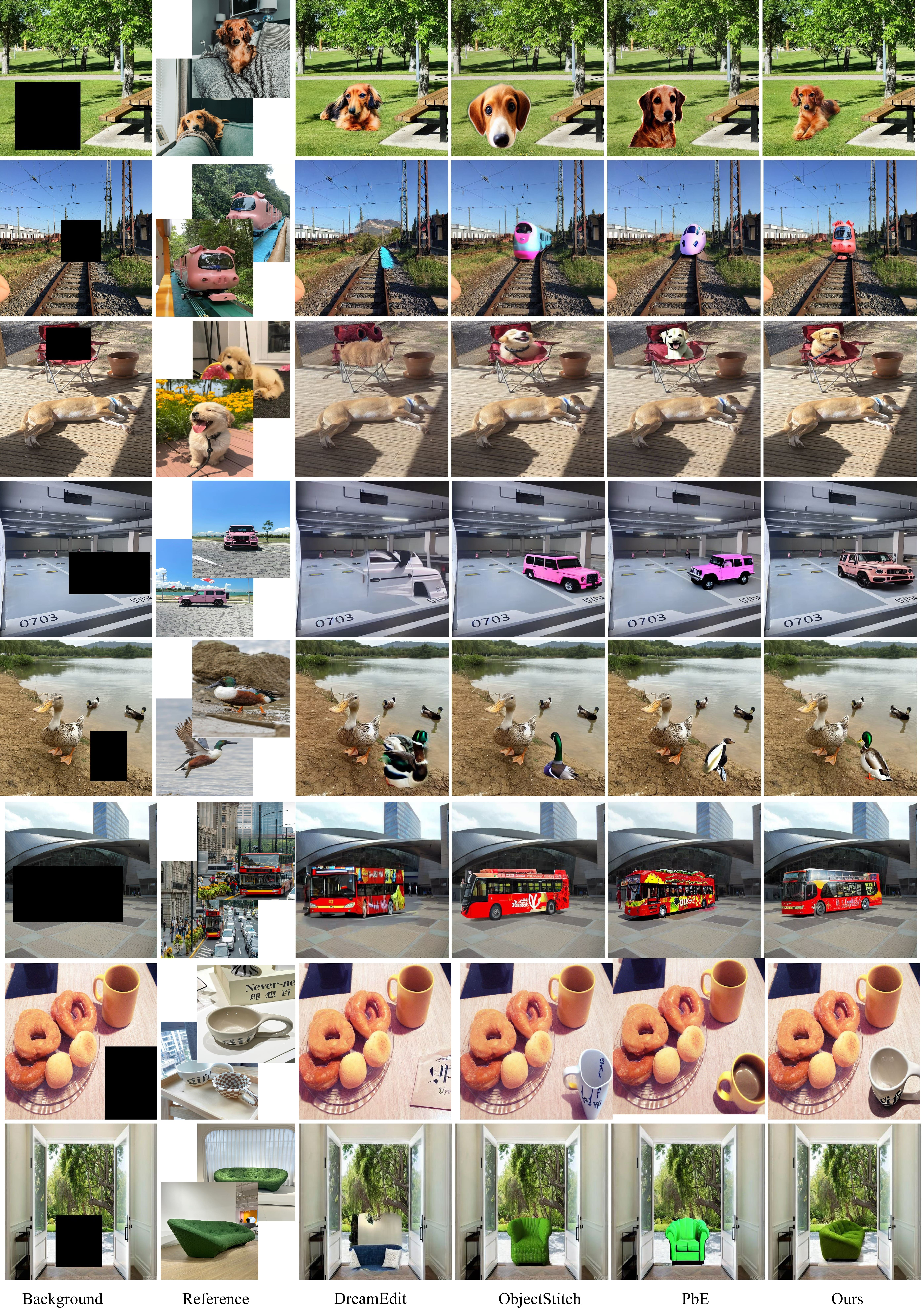}
\caption{Visual results of different methods. In each row, we show the background image with a bounding box, two reference images of foreground object, and the results of DreamEdit \cite{li2023dreamedit}, ObjectStitch \cite{objectstitch}, PbE \cite{paintbyexample}, and our DreamCom. }
\label{fig:baseline1}
\end{figure*}

\begin{figure*}[t]
\centering
\includegraphics[width=0.89\linewidth]{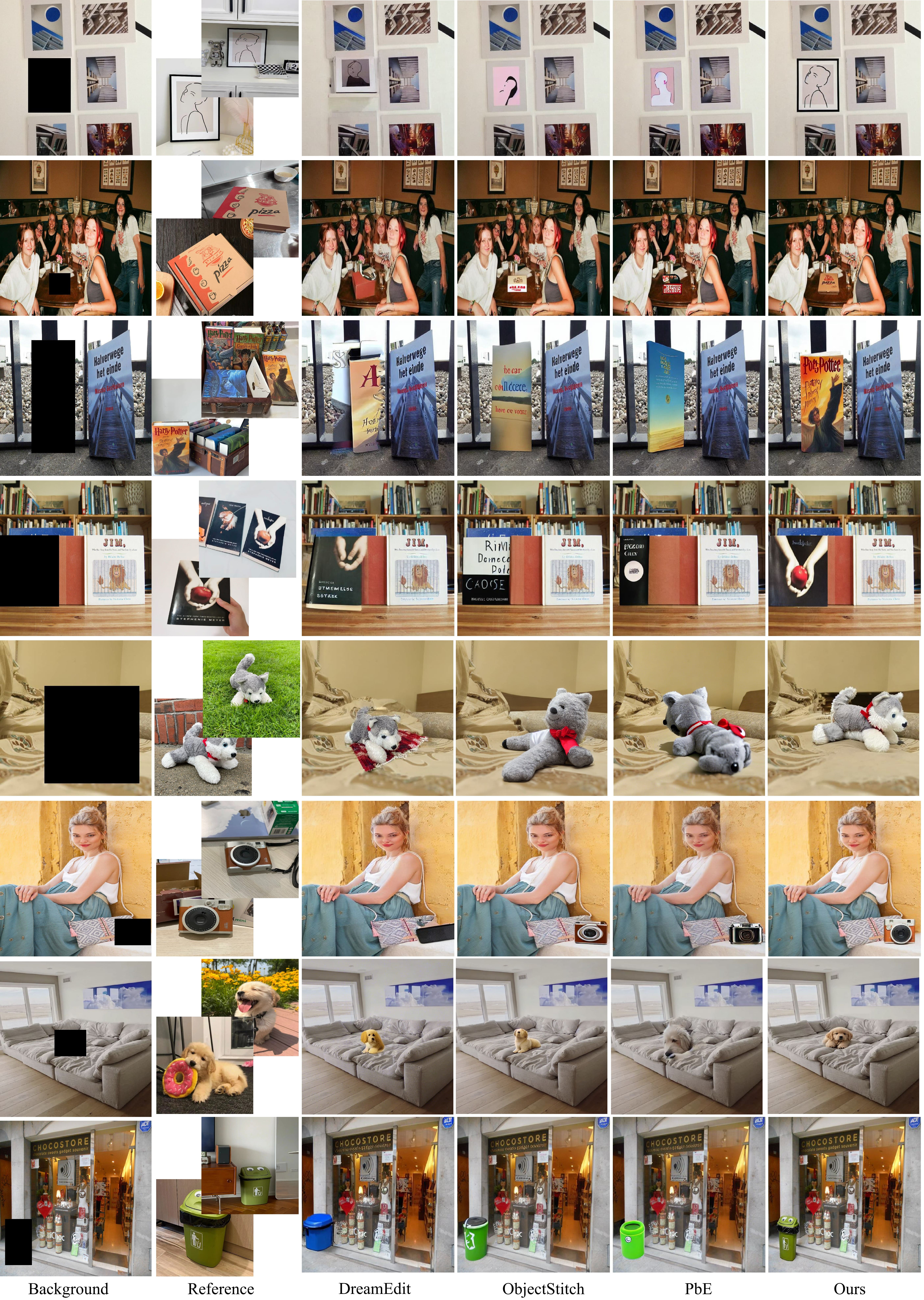}
\caption{Visual results of different methods. In each row, we show the background image with bounding box, two reference images of foreground object, and the results of DreamEdit \cite{li2023dreamedit}, ObjectStitch \cite{objectstitch}, PbE \cite{paintbyexample}, and our DreamCom.}
\label{fig:baseline2}
\end{figure*}

We provide more visual comparison with other baselines.
The results are presented in Fig.~\ref{fig:baseline1} and Fig.~\ref{fig:baseline2}. Recall that we compare with three baselines, two for object-to-object mapping (PbE~\cite{paintbyexample} and ObjectStitch~\cite{objectstitch}), and one for token-to-object mapping (DreamEdit~\cite{li2023dreamedit}).

The issues with PbE~\cite{paintbyexample}  and ObjectStitch~\cite{objectstitch} are evident. Firstly, they could generate the object from the same semantic category of reference object, but the object details between generated object and reference object are dramatically different. 
Secondly, they can only use a single reference image, which severely limits the generation of diverse foreground poses. The foreground pose may not be well adjusted and adapted, when the viewpoints of the background and the reference image have huge gap, resulting in deformed and distorted shapes of generated objects that make the whole image more unrealistic (\emph{e.g.}, row 1-3, 5 in Fig.~\ref{fig:baseline1}).

Due to the strengths of Dreambooth~\cite{dreambooth}, DreamEdit~\cite{li2023dreamedit} can generate objects that resemble the reference objects, but some details may be altered in an unexpected manner. DreamEdit simply combines the foreground and background through the denoising process, leads to insufficient interaction between the foreground and background. When the background is complex, it becomes difficult for the generated foreground to be blended with the background effectively (\emph{e.g.}, row 2-5 in Fig.~\ref{fig:baseline1}). Additionally, DreamEdit~\cite{li2023dreamedit} heavily relies on foreground initialization. Two initialization strategies are provided in \cite{li2023dreamedit}. The first one is GLIGEN~\cite{gligen}, which generates an object similar to the reference object in the background. The second one directly extracts the reference object from the reference image and pastes it on the background. 
By default, we adopt the first initialization strategy for DreamEdit because it can  generate object poses that fit into the semantic context, and reduce the discrepancy between background and foreground, making it easier for the background to generate the correct foreground. The second initialization method is not adopted because it directly copies and pastes the foreground objects, making it difficult to adjust the poses based on the background.  Although the first initialization strategy can help solve the problem of generating reasonable poses, it still struggles when the colors of the generated object and the reference object differ a lot, making the subsequent transformation process very difficult (\emph{e.g.}, row 4, 8 in Fig.~\ref{fig:baseline1}). 
Furthermore, multiple rounds of segmentation and generation do not necessarily lead to positive progress in image quality. For example, if the generated result is poor in the first iteration and the object mask is extracted from the poor result, it may lead to inferior outcomes in the succeeding iterations (\emph{e.g.}, row 2-5, 7 in Fig.~\ref{fig:baseline1}).  

In comparison with previous methods, our method makes significant contributions in two aspects: Firstly, it achieves more seamless integration between the foreground and background,  in terms of poses and lighting (\emph{e.g.}, row 1-8 in Fig.~\ref{fig:baseline1}). Secondly, our method can better preserve the details of foreground appearance and textures (\emph{e.g.}, row 1-5 in Fig.~\ref{fig:baseline2}).

{\small
\bibliographystyle{ieee_fullname}
\bibliography{supp.bbl}
}